\documentclass[10pt,twocolumn,letterpaper]{article}

\usepackage[pagenumbers]{cvpr} 
\usepackage[normalem]{ulem} 
\usepackage{booktabs}
\usepackage{subcaption}
\definecolor{cvprblue}{rgb}{0.21,0.49,0.74}
\usepackage[pagebackref,breaklinks,colorlinks,allcolors=cvprblue]{hyperref}

\def\paperID{5195} 
\def\confName{CVPR}
\def\confYear{2026}

\title{FoSS:
Modeling Long-Range Dependencies and Multimodal Uncertainty in Trajectory Prediction via Fourier–State Space Integration}

\author{Yizhou Huang\textsuperscript{1}\thanks{First author: byyh009@brunel.ac.uk}, Gengze Jiang\textsuperscript{1},
Yihua Cheng\textsuperscript{2}, Kezhi Wang\textsuperscript{1}\thanks{Corresponding author: kezhi.wang@brunel.ac.uk}\\
\textsuperscript{1}Brunel University of London, \textsuperscript{2}University of Birmingham\\
}

\begin{document}
\maketitle
\begin{abstract}
Accurate trajectory prediction is vital for safe autonomous driving, yet existing approaches struggle to balance modeling power and computational efficiency. Attention-based architectures incur quadratic complexity with increasing agents, while recurrent models struggle to capture long-range dependencies and fine-grained local dynamics. 
Building upon this, we present \textbf{FoSS}, a dual-branch framework that unifies frequency-domain reasoning with linear-time sequence modeling. The frequency-domain branch performs a discrete Fourier transform to decompose trajectories into amplitude components encoding global intent and phase components capturing local variations, followed by a progressive helix reordering module that preserves spectral order; two selective state-space submodules, Coarse2Fine-SSM and SpecEvolve-SSM, refine spectral features with $\mathcal{O}(N)$ complexity. In parallel, a time-domain dynamic selective SSM reconstructs self-attention behavior in linear time to retain long-range temporal context. 
A cross-attention layer fuses temporal and spectral representations, while learnable queries generate multiple candidate trajectories, and a weighted fusion head expresses motion uncertainty. Experiments on Argoverse~1 and Argoverse~2 benchmarks demonstrate that \textbf{FoSS} achieves state-of-the-art accuracy while reducing computation by 22.5\% and parameters by over 40\%. Comprehensive ablations confirm the necessity of each component.
\end{abstract}    
\section{Introduction}
Accurate trajectory prediction is essential for safe autonomous driving, particularly in dense multi-agent environments where vehicles and pedestrians interact dynamically \cite{huang2025trajectory, Cheng_2024_CVPR}. Modern prediction systems must simultaneously satisfy three requirements: modeling long-range dependencies across agents, representing multimodal futures to capture uncertainty, and operating within strict real-time constraints. Despite substantial progress, existing approaches~\cite{casas2020implicit,salzmann2020trajectron++,deo2022multimodal,chen2022scept} face inherent trade-offs between accuracy, efficiency, and scalability. Methods relying solely on temporal modeling often conflate global motion patterns with local dynamics, while transformer-based architectures~\cite{yan2023int2,zhou2023query,jia2023hdgt,zhang2023real,seff2023motionlm} achieve high accuracy at the cost of quadratic computational complexity, limiting their deployment in resource-constrained systems.

\begin{figure}[t]
\centering
\includegraphics[width=0.48\textwidth]{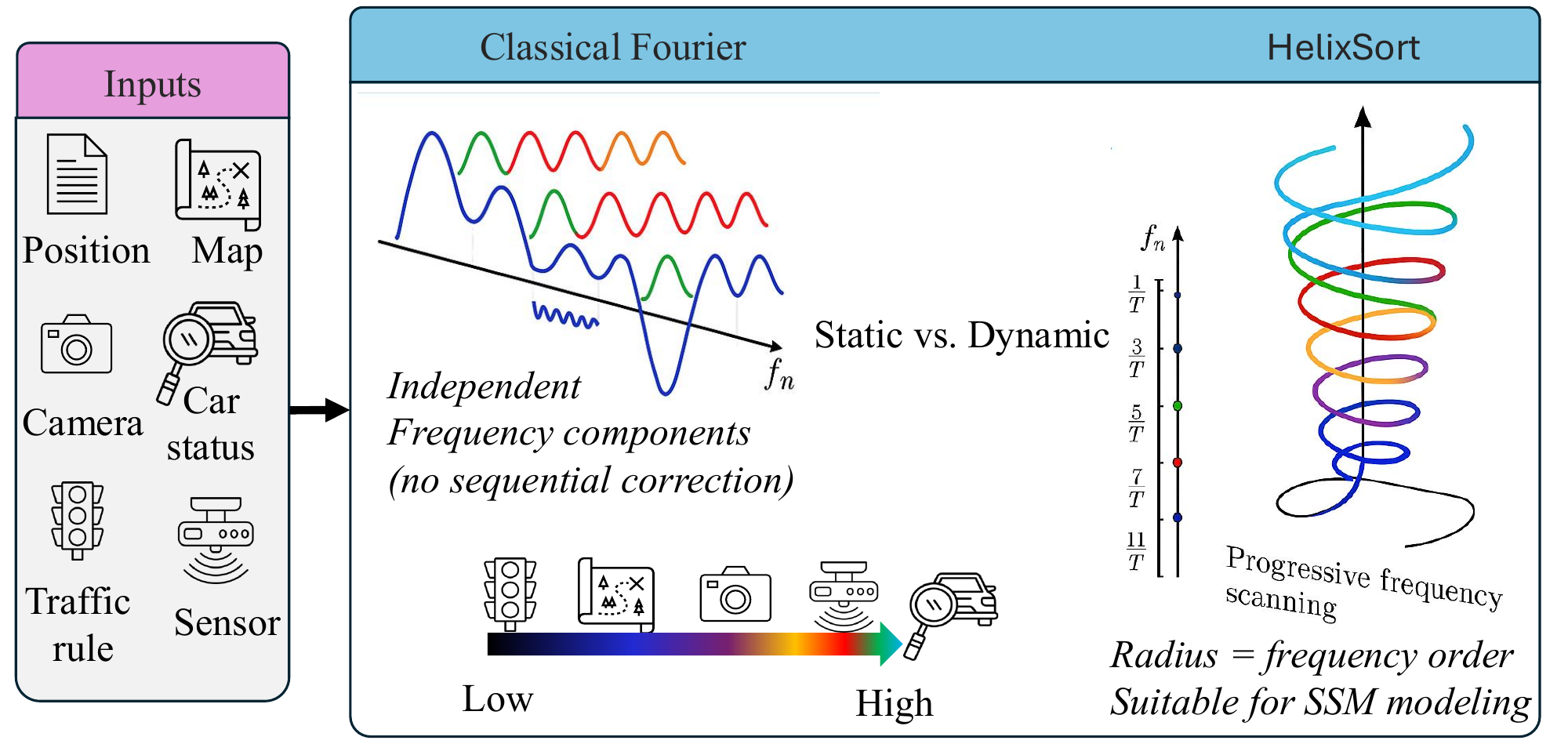} 
\caption{The relationship between input modalities, classical Fourier decomposition, and our HelixSort progressive scanning. Low-frequency inputs such as traffic rules and maps encode global priors, while high-frequency inputs such as vehicle status and position contain fine-grained dynamics. HelixSort progressively scans and aligns these heterogeneous frequency cues, forming an ordered frequency sequence for selective state-space modeling.}
\label{Figure 1}
\end{figure}

In response, we present \textbf{FoSS}, to the best of our knowledge, it is one of the first principled integrations of frequency-domain analysis and linear-complexity sequence modeling in large-scale autonomous driving trajectory prediction. We observe that trajectory signals exhibit complementary structures in the spectral and temporal domains: amplitude spectra encode global motion trends, while phase spectra capture fine-grained temporal variations. However, standard Fourier representations lack ordered frequency semantics, making it difficult for sequence models to process spectral information effectively \cite{wong2022view,cao2021spectral,neumeier2022multidimensional}. Moreover, conventional fusion strategies \cite{sinha2022nl,sharma2021deep} between temporal and spectral features often suffer from scale mismatches that hinder optimization stability. 

Our framework employs a dual-branch architecture to preserve and synergistically combine both representations. The \textit{frequency-domain branch} performs discrete Fourier decomposition followed by a progressive helix reordering module (\textbf{HelixSort}) to impose structural ordering, enabling selective state space modules, namely \textit{Coarse2Fine-SSM} and \textit{SpecEvolve-SSM} to refine spectral features with linear computational complexity. The \textit{time-domain branch} applies an input-dependent selective state space model~\cite{gu2023mamba,gu2022parameterization,smith2024convolutional} to capture long-range dependencies efficiently. A \textit{cross-attention mechanism} fuses the two branches, resolving feature-scale mismatches through normalization and residual connections. Finally, a learnable query-based decoder generates multimodal future trajectories through a weighted candidate fusion strategy.

Extensive experiments on the Argoverse 1~\cite{Argoverse} and Argoverse 2~\cite{TrustButVerify,Argoverse2} benchmarks demonstrate that FourierMamba-TP achieves state-of-the-art accuracy while reducing computational cost and parameter count, validating its scalability for real-world autonomous systems.
Our contributions are summarized as: 1) We propose a dual-branch framework that integrates Fourier-domain decomposition with selective state space modeling, enabling disentangled representation of global and local motion dynamics.
2) We design two selective SSM submodules, Fourier Coarse-to-Fine Frequency (Coarse2Fine-SSM) and frequency channel evolution (SpecEvolve-SSM), that refine spectral features with linear computational complexity.
3) We improve the cross-attention mechanism for stable fusion between temporal and spectral representations, enabling efficient and multimodal trajectory prediction.

\section{Related Work}
\label{gen_inst}
\subsection{Fourier Transform in Vision Tasks}
The Fourier transform is a classical tool for decomposing time- or space-domain signals into frequency components, separating low-frequency structures from high-frequency details and noise.
Such properties have been widely explored in super-resolution~\cite{sinha2022nl,nehete2024fourier} and de-raining tasks~\cite{li2024fouriermamba,sharma2021deep}.
Several recent studies \cite{wong2022view,cao2021spectral,neumeier2022multidimensional} have investigated the integration of Fourier-based representations into trajectory prediction. These methods typically employ spectral decomposition to encode motion patterns in the frequency domain, facilitating the separation of global trends from local variations.
Unlike prior Fourier-based approaches, our proposed Helix sweeping mechanism explicitly decomposes trajectory signals into low-frequency global trends and high-frequency local variations. By transforming these into structured priors for State Space Models (SSMs), we provide essential frequency-domain guidance that complements the long-range dependency modeling of SSMs. This design not only strengthens the representation of complex motion patterns but also successfully extends the Fourier-based reasoning paradigm to the realm of non-linear trajectory prediction.

\subsection{Selective State Space Models (SSMs)}
The selective SSM extends traditional linear SSMs by incorporating input-dependent parameterization, enabling dynamic adjustment of state transitions based on varying inputs \cite{gu2023mamba}. This architecture enhances the model's ability to capture long-range dependencies across sequences \cite{rezaei2024mambalrp}.
Recent studies \cite{wang2023selective,fathullah2023multi,li2024videomamba,oshima2024ssm} have successfully applied selective SSMs to long-range modeling tasks in domains such as video and speech, demonstrating their efficiency and representational capacity.
In autonomous driving scenarios, trajectory data often contain both global motion trends and local dynamic variations. The input-adaptive structure of selective SSMs is well aligned with this dual-level information, allowing efficient modeling of complex spatiotemporal patterns. Temporal-only models struggle to capture both global motion trends and local dynamics.
This work designs a dual-branch network that embeds selective SSMs in the frequency domain, where spatial- and channel-level spectra are deeply refined, yielding richer multimodal representations and more effective feature fusion than existing SSM approaches.

\subsection{Multimodal Motion Prediction Model}
Deep learning–based trajectory prediction has seen significant progress through a variety of architectures, including RNNs, LSTMs, and transformers. Early works such as Social-LSTM and its extensions \cite{li2021lane, rella2021decoder} utilize recurrent structures to model temporal dependencies but often suffer from vanishing gradients when handling long sequences, limiting their effectiveness in complex driving scenarios.
Transformer-based approaches \cite{yan2023int2, zhou2023query, jia2023hdgt} improve interaction modeling through self-attention mechanisms, yet incur quadratic computational complexity with respect to sequence length, making them less suitable for real-time applications involving dense multi-agent interactions.
To address uncertainty in future motion, several studies \cite{huang2025trajectory,Tang_2024_CVPR,zhou2023query,liang2020learning} have proposed candidate-based prediction schemes followed by weighted fusion. While these methods improve multimodal expressiveness, they often rely on heuristics or static fusion weights, which may limit the adaptability of the model under highly dynamic conditions.
\section{Methodology}
\label{headings}
\subsection{Problem Formulation}
Let the historical trajectory sequence be denoted as $X \in \mathbb{R}^{T \times d}$, where $T$ is the number of observed time steps and $d$ is the feature dimension at each step (e.g., 2D or 3D position, velocity, etc.). The goal of trajectory prediction is to estimate the future trajectory $Y \in \mathbb{R}^{T^{\prime} \times d}$ over $T^{\prime}$ future time steps based on the observed sequence $X$. Conventional approaches typically model $X$ directly in the time domain. However, when dealing with complex scenarios and multi-scale information, they often struggle to capture both global motion trends and local dynamic details simultaneously. This method builds upon the standard problem formulation and proposes a dual-branch approach that decomposes the observed sequence $X$ in the frequency domain and integrates it with time-domain information for prediction. This design aims to improve both prediction accuracy and computational efficiency.

\begin{figure*}[t]
\centering
\includegraphics[width=1.0\textwidth]{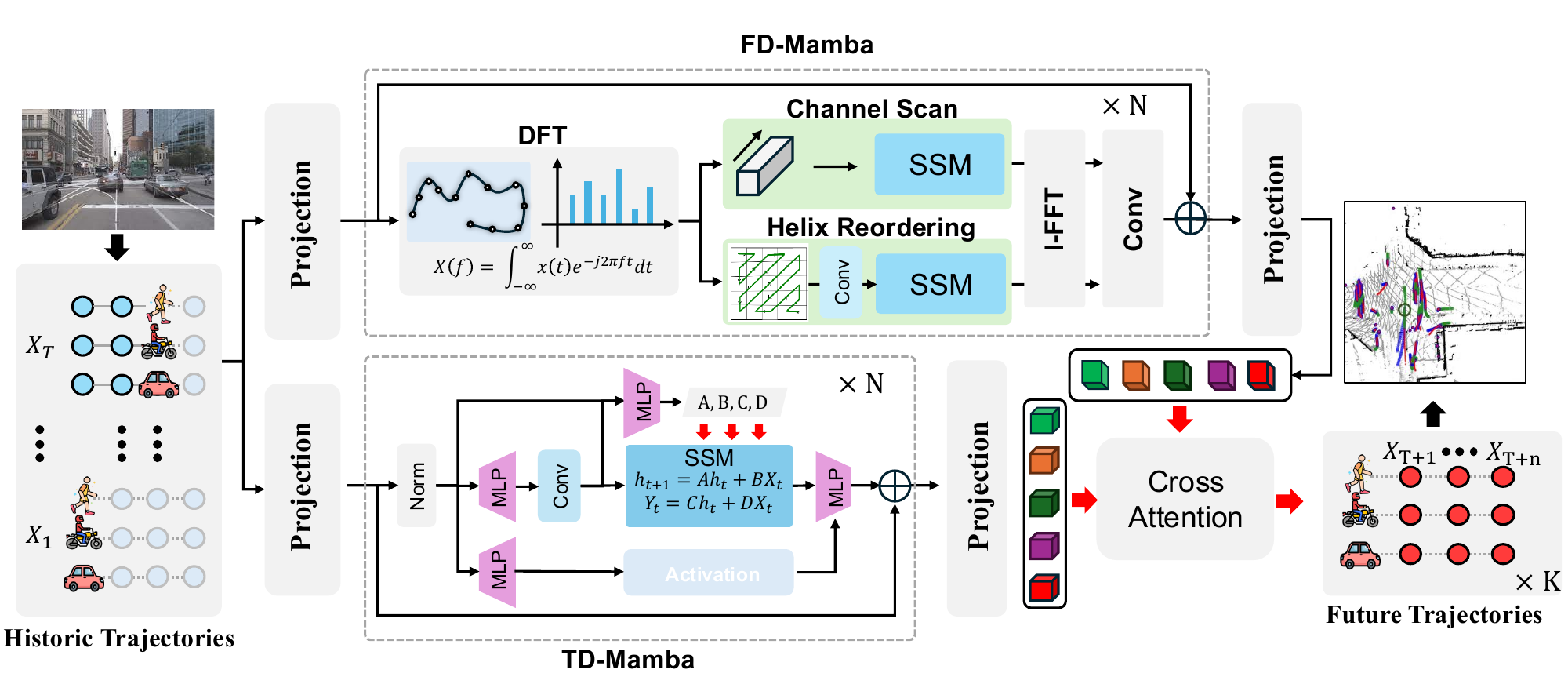} 
\caption{Overview of the proposed framework. Historical trajectories are processed in two parallel branches: (i) frequency branch (FD-Mamba) applies a DFT and reorders the spectrum with a progressive HelixSort in two selective state-space blocks. Coarse2Fine-SSM for spatial interaction and SpecEvolve-SSM for channel evolution; (ii) time branch (TD-Mamba) passes the raw sequence through an input-dependent selective SSM where each state-transition matrix is dynamically generated from the current observation and its local Conv1D features. This design allows the temporal dynamics to adaptively evolve over time, effectively mimicking self-attention behavior with linear complexity.The resulting features are fused via a cross-attention layer, after which a learnable query set decodes $K$ candidate futures whose uncertainty-aware weights yield the final trajectory prediction.}
\label{Figure 2}
\end{figure*}

\subsection{Fourier Transform with SSMs}
In practical applications, the Fourier transform involves several important considerations: \textbf{Normalization factor.} The factor $\frac{1}{\sqrt{T}}$ ensures consistency between the forward and inverse Fourier transforms. It preserves the total energy of the signal across both the time and frequency domains. This property helps maintain numerical stability and facilitates reliable modeling of frequency-domain features. \textbf{Complex-valued representation.} Since $F(X)(\omega) \in \mathbb{C}$ is a complex number, its real part $\Re\{F(X)(\omega)\}$ and imaginary part $\Im\{F(X)(\omega)\}$ correspond to the projections of the original signal onto cosine and sine bases, respectively. By analyzing these two components, the model can effectively identify behavioral patterns associated with different frequencies. This facilitates an explicit decomposition of the signal into global structures and local perturbations. 

The core idea of the \textbf{FoSS} framework is to represent a time-domain signal as a sum of sinusoidal and cosinusoidal components at different frequencies. This decomposition enables the separation of global trends and local variations based on their frequency characteristics. In trajectory modeling tasks, such a representation facilitates the identification of periodic behavioral patterns and fine-grained dynamic changes.
The standard formulation of the Discrete Fourier Transform (DFT)\footnote{For simplicity, we use the term “FFT” throughout this paper to refer broadly to both the Discrete Fourier Transform (DFT) and its fast computational implementation, without distinguishing between their theoretical and algorithmic meanings.} is given by
\begin{equation}
\label{deqn_ex1}
    F(X)(\omega)=\frac{1}{\sqrt{T}} \sum_{t=0}^{T-1} X(t) e^{-j 2 \pi \frac{t \omega}{T}}, \omega \in\{0,1, \ldots, T-1\},
\end{equation}
where $X (t)\in \mathbb{R}^{T \times d}$ denotes the trajectory feature at time step $t$. $\omega$ is the frequency index, and $j$ is the imaginary unit.

After obtaining the Fourier transform result $F(X)(\omega) \in \mathbb{C}$, we further decompose it into amplitude $A(X)(\omega)$ and phase $P(X)(\omega)$ components to separately model the global trends and local dynamic behaviors in the trajectory. Specifically, for each frequency index $\omega$, 
\begin{equation}
\label{deqn_ex2}
\begin{aligned}
A(X)(\omega) &= \sqrt{\Re\{F(X)(\omega)\}^{2}+\Im\{F(X)(\omega)\}^{2}},\\
P(X)(\omega) &= \arctan\left(\frac{\Im\{F(X)(\omega)\}}{\Re\{F(X)(\omega)\}}\right),
\end{aligned}
\end{equation}
the $A(X)$ captures of the energy distribution of the signal across different frequencies, which is typically associated with global motion trends and periodic behavioral patterns. The $P(X)$ encodes the temporal alignment of individual frequency components, which can reveal fine-grained temporal dynamics and local perturbations in the trajectory.

Conventional Fourier-transform outputs do not preserve a continuous ordering from low to high frequency, 
because DFT symmetry, \ie, $\omega$ and $T-\omega$ correspond to the same physical frequency.
To address this, we introduce HelixSort$(\cdot)$, a helix reordering module inspired by JPEG's zigzag encoding~\cite{jpeg1994standard}. 
For multi-dimensional trajectories ($d > 1$), HelixSort operates independently along each feature dimension $k \in \{1, \dots, d\}$. The detailed steps are as follows:

\begin{enumerate}
    \item Reshape the 1D DFT coefficients $F^{(k)} \in \mathbb{C}^{T}$ into a 2D grid $\mathcal{F}^{(k)} \in \mathbb{C}^{\sqrt{T} \times \sqrt{T}}$, zero-padded to the nearest square dimension to preserve spectral information.
    \item Starting from the DC component at the spectral center $(u_0, v_0)$, traverse $\mathcal{F}^{(k)}$ in an outward spiral order. Coefficients are sorted by ascending spectral radius $r = \sqrt{(u - u_0)^2 + (v - v_0)^2}$, with ties broken by row-major order.
    \item Record the index mapping $\pi^{(k)}$ to produce a reordered 1D sequence $\widehat{F}^{(k)} \in \mathbb{C}^{T}$.
\end{enumerate}

This yields a non-decreasing radius sequence, \ie, $\forall i < j, \; r_i \leq r_j$, 
which concentrates low-frequency components (global trends) at the sequence start 
and high-frequency components (local dynamics) toward the end. 
This ordered representation enables sequential state-space models to process spectral information in a coarse-to-fine manner.

\textbf{Compatibility with SSMs.} Applying SSMs directly to standard DFT outputs is suboptimal, the interleaved low/high-frequency coefficients force the model to constantly switch between global and local reasoning, disrupting the state evolution. HelixSort resolves this by providing an ordered spectral sequence. State space models (SSMs) are naturally suited to long-range modeling over one-dimensional sequences. By interpreting the reordered sequence $\widehat{F}$ from HelixSort as an ``artificial temporal axis'', we define the recurrence:

\begin{equation}
h_{s+1} = A_s h_s + B_s \widehat{F}(s), \quad Y_{\mathrm{freq}}(s) = C_s h_s + D_s \widehat{F}(s),
\end{equation}

where the index $s$ reflects the frequency rank (via spectral radius $r_s$) rather than real time. As low-frequency components are processed first, the SSM accumulates global trends in its state. When high-frequency components arrive later, the model has already built sufficient context to modulate short-term details—a process aligned with the common coarse-to-fine reasoning strategy. Moreover, the monotonicity of HelixSort allows the parameters $A_s$, $B_s$ to specialize in low-frequency dynamics, while high-frequency inputs are selectively gated to suppress noise. This separation significantly reduces the conditioning number of the state transition matrix $A_s$, leading to more stable gradient propagation during training.

\textbf{Complexity Analysis.} HelixSort involves one FFT operation ($\mathcal{O}(T \log T)$) and index reordering ($\mathcal{O}(T)$). It stores a $T$-dimensional index map $\pi$, introducing no learnable parameters and negligible overhead: $0.08\pm 0.015\%$ total FLOPs increase and $<$0.25MB memory (measured on Argoverse 2, $T=50$).

\subsection{Dual-Branch Network Architecture}
Considering that modeling trajectory data using only time-domain or frequency-domain methods may be insufficient to fully capture its multi-scale characteristics, we propose a dual-branch network architecture. As illustrated in Figure \ref{Figure 1}, this architecture consists of a time-domain branch and a frequency-domain branch; it respectively models and interprets trajectories from the perspectives of temporal evolution and frequency structure.

\textbf{TD-Mamba: Selective state space modeling in time-domain branch.} In the time domain branch, the input trajectory data $X \in \mathbb{R}^{T \times d}$ is directly used as input. A selective SSM is employed to extract dynamic evolution features of the target's motion process from it and capture long-range dependencies. In traditional linear SSMs, state updates are typically performed using the following formula:
\begin{equation}
\label{deqn_ex4}
    h(t+1) =A h(t)+B X(t), \
    Y_{\text {time }}(t) =C h(t)+D X(t),
\end{equation}
where $h(t) \in \mathbb{R}^{n}$ represents the hidden state at time $t$, and $A \in \mathbb{R}^{n \times n}, B \in \mathbb{R}^{n \times d}, C \in \mathbb{R}^{p \times n}, D \in \mathbb{R}^{p \times d}$ are the state transition, input, output, and feedforward matrices, respectively. However, for complex trajectory data, this fixed-parameter approach often struggles to fully suppress noise and capture crucial dynamic information. Therefore, a selective state space model is introduced in our model, dynamically adjusting each parameter matrix based on the current input $X(t)$ and its local convolutional features $\tilde{X}(t)=\operatorname{Conv1D}(X(t))$. Specifically, the matrix parameters are defined by
\begin{equation}
\begin{alignedat}{2}
A_t &= f_A(X(t), \tilde{X}(t)), \quad &B_t &= f_B(X(t), \tilde{X}(t)), \\
C_t &= f_C(X(t), \tilde{X}(t)), \quad &D_t &= f_D(X(t), \tilde{X}(t)).
\end{alignedat}
\end{equation}
Where $f_{A}, f_{B}, f_{C}, f_{D}$ all are implemented by lightweight Multilayer Perceptrons (MLPs), used to generate the corresponding parameters based on the current input information. Therefore, the state update formula is rewritten as:
\begin{equation}
\begin{alignedat}{2}
h(t+1)         &= A_{t} h(t) + B_{t} X(t), \quad
&Y_{\text{time}}(t) &= C_{t} h(t) + D_{t} X(t).
\end{alignedat}
\end{equation}
This input-dependent parameterization allows the state update process to automatically and selectively amplify key motion patterns while suppressing detrimental noise interference based on the input at different time steps. Furthermore, during the update process, the input is preprocessed using 1D convolution (Conv1D) to extract local short-term features. The output of this convolution $\tilde{X}(t)$ is integrated into the parameter generation functions ($f_{A}, f_{B}, f_{C}, f_{D}$), thereby enhancing the model's sensitivity to local dynamic changes.
To ensure numerical stability, the hidden state at each time step, after undergoing SiLU activation, is also subjected to layer normalization, specifically,
\begin{equation}
    \hat{h}(t+1) = \text{LayerNorm}(\text{SiLU}(h(t+1))),
\end{equation}
after undergoing the dynamic state update and normalization operations described above, the entire time domain branch outputs a series of features $Y_{\text {time }} \in \mathbb{R}^{T \times p}$. This feature not only reflects the continuous dynamics of the motion sequence, but also possesses good anti-interference capabilities, providing effective temporal priors for frequency-domain feature fusion and candidate trajectory generation.

\textbf{FD-Mamba: Spectrum reconstruction and sequential modeling in frequency-domain branch.} To effectively capture the global motion trends and local dynamic information of trajectory data, the data $X$ is first transformed to the frequency domain using Equation (\ref{deqn_ex1}). Subsequently, the amplitude $A(X)$ and phase $P(X)$ are then calculated separately via Equation (\ref{deqn_ex2}).

Since the frequency components output by the traditional Fourier transform do not directly present a low-frequency to high-frequency order, the progressive helix reordering module HelixSort$(\cdot)$ addresses this limitation. This module first rearranges the amplitude $A(X)$ and phase $P(X)$ based on the distance of each frequency component from the spectrum center, thereby concentrating low-frequency information in the initial part of the sequence and ordering high-frequency information subsequently. The rearranged amplitude and phase are denoted by 
\begin{equation}
\label{deqn_ex3}
\begin{aligned}
A^{\prime}(X)(\omega) &= \text{HelixSort}(A(X)(\omega)),\\
P^{\prime}(X)(\omega) &= \text{HelixSort}(P(X)(\omega)).
\end{aligned}
\end{equation}

Building upon this, we further incorporate two sub-modules to deeply model frequency domain features from spatial and channel perspectives: (a) the Fourier Spatial Interaction State Space Module (Coarse2Fine-SSM) and (b) the Fourier Channel Evolution State Space Module (SpecEvolve-SSM). Both submodules operate in parallel on the shared input $F_{in}$ (obtained as $F_l$ after LayerNorm transformation), modeling spatial and channel spectral dependencies, respectively.

Firstly, in Coarse2Fine-SSM, $F_{in}$ undergoes FFT to convert to the frequency domain, producing the corresponding amplitude spectrum $A(F_l)$ and phase spectrum $P(F_l)$. Subsequently, the progressive helix reordering module separately rearranges these two parts: 
\begin{equation}
\begin{alignedat}{2}
A^{\prime}(F_{l}) &= \text{HelixSort}(A(F_{l})), \quad
&P^{\prime}(F_{l}) &= \text{HelixSort}(P(F_{l})).
\end{alignedat}
\end{equation}
To further refine the frequency-domain information, the module sequentially processes the rearranged sequences using Depthwise Separable Convolution (DWConv), SiLU activation, a Selective SSM, and LayerNorm. Finally, restoring these processed features back to the time domain via inverse FFT, followed by element-wise multiplication with the original SiLU-activated $F_l$, forms the spatially interactive representation: 
\begin{equation}
    F_{f} = \text{iFFT}\big(A^{\prime}(F_{l}), P^{\prime}(F_{l})\big) \odot \operatorname{SiLU}(F_{l}).
\end{equation}

Secondly, the SpecEvolve-SSM module processes spectral information along the channel dimension. Specifically, the input feature $F_{in} \in \mathbb{R}^{H\times W \times C}$ undergoes global average pooling to produce $F_g \in \mathbb{R}^{1\times 1 \times C}$. Fourier Transform is applied to $F_g$, producing channel-wise spectrum $F(F_g)(z)$ with amplitude $A(F_g)(z)$ and phase $P(F_g)(z)$. To exploit inter-channel correlations, a one-dimensional progressive helix reordering (Channel Scan) reorders coefficients by ascending spectral magnitude $|F(F_g)(z)|$. The reordered spectra $A(F_g)'(z)$ and $P(F_g)'(z)$ are restored via inverse FFT, yielding processed channel features. These are element-wise multiplied with $F_g$, activated by SiLU, to produce the evolved channel feature:
\begin{equation}
    F_{a} = \text{iFFT}\big(A(F_{g})'(z), P(F_{g})'(z)\big) \odot \operatorname{SiLU}(F_{g}).
\end{equation}
The evolved feature $F_a$ is fused with $F_{in}$ via element-wise gating to obtain the enhanced representation:
\begin{equation}
    F_{\text{enhance}} = F_{a} \odot F_{in}.
\end{equation}

Finally, the output $F_f$ from the Coarse2Fine-SSM and the $F_{\text{enhance}}$ from the SpecEvolve-SSM are concatenated along the channel dimension and passed through a linear projection layer to align their scales and distributions, yielding the final frequency-domain representation $F_{\mathrm{freq}}$.
This preserves the spatial-interaction cues extracted by Coarse2Fine-SSM and fuses the channel-evolution features produced by SpecEvolve-SSM. Consequently, $F_{\mathrm{freq}}(t)$ captures both global motion trends and local dynamic details, offering a strong frequency-domain prior for subsequent feature fusion and trajectory generation. 

\subsection{Candidate Trajectory Generation and Loss Function}
This module introduces the cross-attention–based candidate-trajectory generation process. It directly derives salient information from the features that combine temporal- and frequency-domain cues, yielding future-trajectory predictions.
Concretely, the dual-branch fused feature $Z \in \mathbb{R}^{T\times d}$ interacts with a set of learnable query vectors $Q \in\mathbb{R}^{K\times d}$,  where $T$ is the sequence length, $d$ is the feature dimension, and \textbf{K} is the number of candidate trajectories. The queries are initialized, whereas the fused feature $Z$ is used as both keys and values, i.e., $K$ = $Z$ and $V$ = $Z$, with $d_k$ denoting the key dimension. This cross-attention mechanism enables each query to aggregate dynamic and structural information relevant to the target motion, producing contextual representations for the candidate trajectories, denoted as: $Z_{atten} \in \mathbb{R}^{K\times d}$. After obtaining the attention output, the feature tensor $Z_{atten}$ is passed into an MLP to apply nonlinear transformations to project the features into the parameter space of candidate trajectories, denoted as: $\hat{Y}_{k}=\operatorname{MLP}\left(Z_{\text {attn }}(k)\right), \quad k=1,2, \ldots, \textbf{K} $.

To jointly constrain prediction quality in both temporal and frequency domains, and to optimize candidate trajectory generation and fusion, this work proposes a unified loss function. First, in the temporal domain, an $\mathcal{L}1$ loss is used to directly measure the difference between the final predicted trajectory $\hat{Y}$ and ground truth $Y$. 
$\mathcal{L}_{\text {time }}=\left\|\hat{Y}_{\text {final }}-Y\right\|_{1}$.
Secondly, to ensure consistency in the frequency domain, both $\hat{Y}$ and $Y$ are transformed using the Fourier transform. An $\mathcal{L}1$ loss is then computed between their frequency-domain representations. $\mathcal{L}_{\text {freq}}=\left\|F(\hat{Y}_{\text {final }})-F(Y)\right\|_{1}$.
And finally, the total loss is defined as:
\begin{equation}
      \mathcal{L}_{\text {total }}=  \mathcal{L}_{\text {time }} +  \lambda \mathcal{L}_{\text {freq }}.
\end{equation}

\section{Experiments}
\subsection{Implementation Details}
All experiments in this work are implemented using the PyTorch Lightning framework. For accuracy comparison, training and evaluation are conducted on an NVIDIA A100 GPU. During training, the batch size is set to 128, and the initial learning rate is 0.001.
The Adam optimizer is used for weight updates, and the model is trained for a total of 50 epochs.
A dynamic learning rate schedule is applied based on validation performance. The learning rate is reduced to 10\% of its original value if the validation metric does not improve for 5 consecutive epochs. Inference latency and efficiency are measured on an NVIDIA RTX 3090 GPU, which closely approximates the runtime characteristics of embedded devices such as the NVIDIA Jetson Orin.  Although RTX 3090 provides higher raw throughput, its shared Ampere architecture with the Jetson Orin allows the results to serve as a reliable proxy for realistic edge deployment conditions.
\subsection{Datasets and Evaluation Metrics}
This study is validated on two large-scale datasets. The Argoverse 1 dataset \cite{Argoverse}, each sample provides 2 seconds of observed trajectory, used to predict the next 3 seconds.
The Argoverse 2 dataset \cite{TrustButVerify,Argoverse2} offers 5 seconds of history for predicting the following 6 seconds, with more complex scene dynamics. Standard trajectory prediction metrics are used for evaluation.
These include the minimum ADE over all time steps (minADE$_K$) and the minimum FDE at the last time step (minFDE$_K$). In addition, the miss rate (MR$_K$) measures the proportion of predicted trajectories whose final point deviates by more than 2 meters. The Brier score variant (b-minFDE$_K$) is also used to assess the calibration of confidence in predictions.
The value of $K$ is set to 1 and 6 following the official prediction protocol.

\begin{table*}[t]
\captionsetup{skip=3pt}
\centering
\caption{Comparison results on the Argoverse 2 dataset \cite{TrustButVerify,Argoverse2}. For each metric, the best result is in \textbf{bold} and \uline{underline} represents the second best. Notably, the reported results reflect the performance of backbone networks only, excluding improvements from engineering techniques such as ensembling.}
\renewcommand\arraystretch{1.1}
\fontsize{9pt}{9pt}\selectfont
\begin{tabular}{l|cccc|c}
\bottomrule[1.5pt]  
Method & b-minFDE $\downarrow$ & minADE $\downarrow$ & minFDE $\downarrow$ & MR $\downarrow$ &\#Params (M)$\downarrow$ \\
\midrule
LaneGCN \cite{liang2020learning} & 2.06 & 0.87 & 1.36 & 0.16 & - \\
DenseTNT \cite{gu2021densetnt} & 1.98 & 0.88 & 1.28 & 0.13 & - \\
BANet \cite{wang2023technicalreportargoversechallenges} & 1.92 & 0.71 & 1.36 & 0.19 & 9.49 \\
SceneTransformer \cite{ngiam2021scene} & 1.89 & 0.80 & 1.23 & 0.13 & 15.30 \\
GOHOME \cite{gilles2022gohome} & 1.86 & 0.89 & 1.29 & 0.08 & - \\
HiVT \cite{zhou2022hivt} & 1.84 & 0.77 & 1.17 & 0.13 & - \\
MultiPath++ \cite{varadarajan2022multipath++} & 1.79 & 0.79 & 1.21 & 0.13 & - \\
PAGA \cite{wolf2019paga} & 1.76 & 0.80 & 1.21 & \textbf{0.11} & - \\
Wayformer \cite{nayakanti2023wayformer} & \uline{1.74} & 0.77 & \uline{1.16} & \uline{0.12} & - \\
QCNet \cite{zhou2023query} & 1.91 & 0.65 & 1.29 & 0.16 & 7.71 \\
DeMo \cite{zhang2024demo} & 1.84 & 0.63 & 1.17 & 0.13 & \uline{5.92} \\
\midrule
\textbf{FD-Mamba$+$Transformer} & 1.77 & 0.75 & 1.21 & 0.14 & 21.83 \\
\textbf{FD-Mamba$+$LSTM} & 1.91& 0.85 & 1.28 & 0.13 & 7.89 \\
\textbf{FoSS (Ours)} & \textbf{1.69} &\textbf{ 0.61} & \textbf{1.07} & \textbf{0.11} & \textbf{4.18} \\
\bottomrule[1.5pt]  
\end{tabular}
\label{tab:1}
\end{table*}

\begin{table*}[t]
\centering
\begin{minipage}[t]{0.52\linewidth}
\vspace{0pt}\centering
\scriptsize
\caption{Comparison results on the Argoverse 1 dataset \cite{Argoverse}. 
The best result is in \textbf{bold} and \uline{underline} represents the second best. 
Performance evaluated on backbone networks without ensembling.}
\setlength{\tabcolsep}{3pt} 
\resizebox{\linewidth}{!}{ 
\begin{tabular}{l|ccc}
\toprule[1.5pt]
Method & minADE$_{1}$ $\downarrow$ & minFDE$_{1}$ $\downarrow$ & MR$_{1}$ $\downarrow$ \\
\midrule
LaneGCN \cite{liang2020learning}           & 1.92 & 2.30 & 0.18 \\
DenseTNT  \cite{gu2021densetnt}            & 1.89 & 2.20 & 0.16 \\
SceneTransformer \cite{ngiam2021scene}     & 1.89 & 2.13 & 0.13 \\
HiVT  \cite{zhou2022hivt}                  & 1.84 & 2.05 & 0.13 \\
MultiPath++  \cite{varadarajan2022multipath++} & 1.79 & 2.08 & 0.13 \\
PAGA  \cite{wolf2019paga}                  & 1.76 & 2.08 & \textbf{0.11} \\
Wayformer  \cite{nayakanti2023wayformer}   & 1.74 & \textbf{2.02} & \uline{0.12} \\
QCNet \cite{zhou2023query} & 1.76 & 2.08 & \textbf{0.11}\\
DeMo \cite{zhang2024demo} & \textbf{1.65} & 3.89 & 0.14\\
\midrule
\textbf{FD-Mamba$+$Transformer} & 1.81 & 2.14 & 0.14 \\
\textbf{FD-Mamba$+$LSTM} & 1.88 & 2.27 & 0.15 \\
\textbf{FoSS (Ours)} & \uline{1.67} & \uline{2.05} & \textbf{0.11} \\
\bottomrule[1.5pt]
\end{tabular}
\label{tab:argo1}
}
\end{minipage}
\hfill
\begin{minipage}[t]{0.45\linewidth}
\vspace{0pt}\centering
\scriptsize
\caption{Ablation study on the Argoverse 2 dataset \cite{TrustButVerify,Argoverse2}. 
F.D.B. refers to the frequency-domain branch; P.H.R.M. is the progressive helix reordering module; 
Fourier S.S.M. indicates the selective state-space model in both Coarse2Fine-SSM and SpecEvolve-SSM submodules; 
Replace Concat$+$MLP  is the replacement of cross-attention in time domain for feature fusion. 
The general compatibility of the frequency branch with other temporal backbones (e.g., LSTM and Transformer) 
is further validated in Table \ref{tab:1}, demonstrating FD-Mamba's plug-and-play performance across different architectures.}
\setlength{\tabcolsep}{4pt}
\resizebox{\linewidth}{!}{
\begin{tabular}{lccc}
\toprule[1.5pt]
Variant & minADE $\downarrow$ & minFDE $\downarrow$ & MR $\downarrow$ \\
\midrule
$w/o$ F.D.B.        & 0.71 & 1.36 & 0.17 \\
$w/o$ P.H.R.M.      & 0.69 & 1.32 & 0.16 \\
$w/o$ Fourier S.S.M.& 0.70 & 1.35 & 0.17 \\
Replace Concat$+$MLP        & 0.69 & 1.33 & 0.16 \\
\addlinespace[2pt]
\textbf{Full model} & \textbf{0.65} & \textbf{1.29} & \textbf{0.15} \\
\bottomrule[1.5pt]
\end{tabular}
}
\label{tab:argo2}
\end{minipage}
\end{table*}

\subsection{Quantitative Experiments}
The comparison includes several state-of-the-art models.
All baselines are evaluated using their publicly released code and official hyperparameter settings or reproduced based on their original papers.

\textbf{Result on Argoverse 2.} 
On the 6-second prediction task of Argoverse 2 (Table~\ref{tab:1}), \textbf{FoSS (Ours)} achieves the best overall performance. 
It records a b-minFDE$_6$ (1.69), improving upon DenseTNT (1.98) by \textbf{14.6\%}, and a minADE$_6$ (0.61), outperforming SceneTransformer (0.75) by \textbf{18.7\%}. 
Its minFDE$_6$ (1.07) shows an \textbf{11.6\%} improvement over the same baseline, and the MR$_6$ (0.11) improves from 0.14 by \textbf{21.4\%}. 
These results confirm that the proposed frequency decomposition and selective SSM substantially enhance long-horizon trajectory accuracy.

\textbf{Result on Argoverse 1.} 
On the 3-second prediction task of Argoverse 1 (Table~\ref{tab:argo1}), \textbf{FoSS (Ours)} achieves a minADE$_1$ (1.67), improving over LaneGCN (1.92) by \textbf{13.0\%}, and a minFDE$_1$ (2.05), comparable to Wayformer (2.02, $\sim$1.5\% difference). 
Its MR$_1$ (0.11) matches the best existing results, indicating strong reliability in short-horizon prediction. 
Overall, the method demonstrates consistent improvements across both datasets with a favorable trade-off between accuracy and efficiency.

\begin{figure*}[t]
\centering
\includegraphics[width=1.0\textwidth]{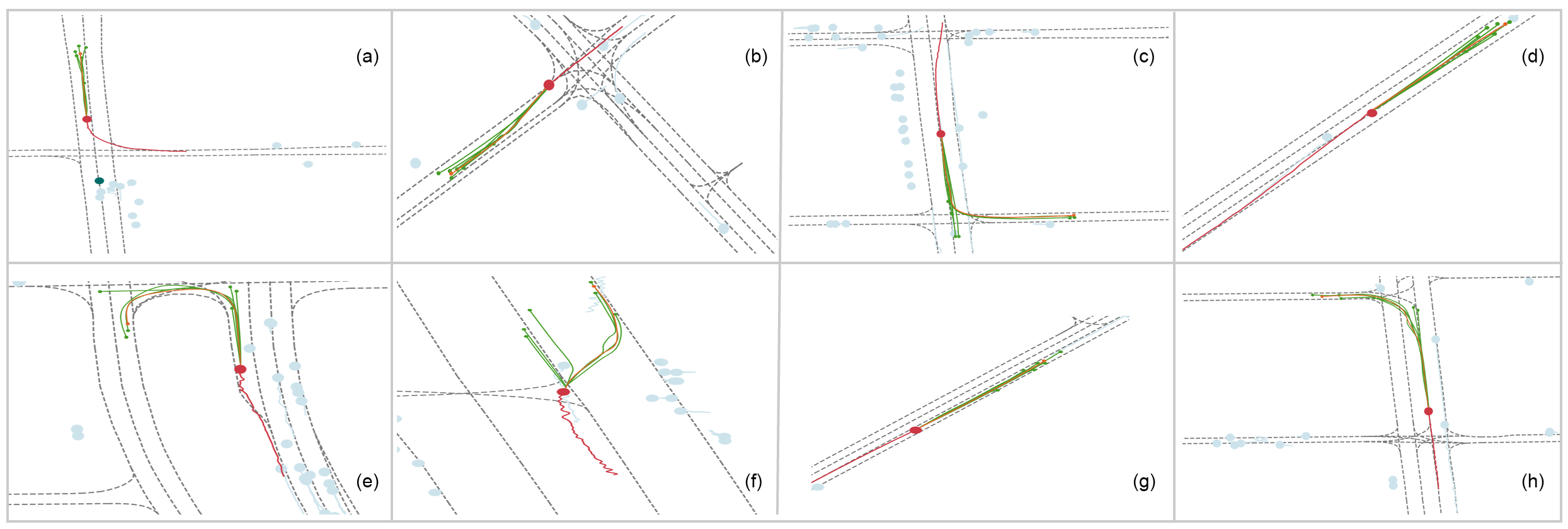} 
\caption{
Qualitative results on Argoverse~2 \cite{TrustButVerify,Argoverse2}, covering U-turn (e), lane change (b,f), turning (a,c,h), and straight-driving scenes (d,g). 
Red: past trajectories; orange: ground truth; green: predicted trajectories. 
The model generates smooth and diverse trajectories aligned with road geometry, with slight jitter (f) in frequent lane-change cases, possibly due to transient high-frequency motion cues.
}
\label{Figure 4}
\end{figure*}

\begin{table*}[t] 
  \centering
  \captionsetup{skip=3pt}
  \renewcommand\arraystretch{1.1}
  \fontsize{9pt}{9pt}\selectfont

  \begin{minipage}[t]{0.48\linewidth}
    \vspace{0pt}\centering
    \begin{tabular}{lc}
      \toprule[1.5pt]
      Method & Online inference (ms) $\downarrow$ \\
      \midrule
      HiVT \cite{zhou2022hivt}               & $82 \pm 1$ \\
      SceneTransformer \cite{ngiam2021scene}    & $76 \pm 1$ \\
      QCNet \cite{zhou2023query}& $71 \pm 1$ \\
      MultiPath$++$ \cite{varadarajan2022multipath++}     & $70 \pm 1$ \\
      \textbf{FoSS (Ours)} & $\mathbf{64 \pm 1}$ \\
      \bottomrule[1.5pt]
    \end{tabular}
    \captionof{table}{Inference latency is evaluated on an NVIDIA RTX 3090 GPU, offering computational characteristics similar to the NVIDIA Jetson Orin developer kit. Performence lower is better.}
    \label{tab:latency}
  \end{minipage}
  \hfill
  \begin{minipage}[t]{0.48\linewidth}
    \vspace{0pt}\centering
    \begin{tabular}{lc}
      \toprule[1.5pt]
      Method & FLOPs (G) $\downarrow$ \\
      \midrule
      GANet \cite{wang2023ganet} & 15.79 \\
      BANet \cite{wang2023technicalreportargoversechallenges} & \textbf{11.93} \\
      QCNet \cite{zhou2023query} & 45.3 \\
      Trajectory Mamba \cite{huang2025trajectory} & 27.3 \\
      \addlinespace[2pt]
      FoSS (Ours) & 22.1 \\
      \bottomrule[1.5pt]
    \end{tabular}
    \captionof{table}{Computational efficiency on Argoverse~2 \cite{Argoverse2,TrustButVerify}. 
    All experiments use an NVIDIA 3090 GPU.}
    \label{tab:flops}
  \end{minipage}
\end{table*}

\textbf{Compatibility with Other Temporal Backbones.} 
We further verify the \emph{plug-and-play} property of the frequency branch\footnote{The hybrid variants are reported only to demonstrate FD-Mamba’s adaptability across architectures, not to claim additional performance gains.}. 
\textbf{FD-Mamba} is combined with two temporal encoders (Transformer and LSTM) by replacing the time-domain branch while keeping all other settings identical. 
These variants fine-tune the original backbones without any structural change. 
Empirically, \textbf{FD-Mamba} aligns naturally with Transformer architectures, as spectral priors complement token-wise aggregation, while our full model remains specifically optimized for SSMs with dynamic parameter updates and coarse-to-fine frequency routing.  

\textbf{Efficiency and Inference Latency.} 
Despite its high accuracy, \textbf{FoSS (Ours)} contains only \textbf{4.18M} parameters, the smallest among all compared models. 
As shown in Table~\ref{tab:latency}, it achieves an average inference time of \textbf{64\,ms}, faster than HiVT (82\,ms), SceneTransformer (76\,ms), QCNet (71\,ms), and MultiPath++ (70\,ms). 
It also requires only \textbf{22.1\,G} FLOPs, a \textbf{51\%} and \textbf{19\%} reduction compared with QCNet (45.3\,G) and Trajectory Mamba (27.3\,G). 
GANet and BANet use CNN- and graph-based encoders derived from LaneGCN; although their FLOPs are lower (15.79 G and 11.93 G), their accuracy drops notably. 
Overall, the dual-branch design achieves a superior trade-off among accuracy, latency, and computational cost.

\subsection{Qualitative Analysis}
Figure~\ref{Figure 4} illustrates representative prediction cases from the Argoverse~2 dataset, covering diverse driving scenarios such as U-turns, lane changes, turns, and straight driving. 
In each example, the red lines denote historical trajectories, orange lines represent ground-truth futures, and green lines are the predictions of \textbf{FoSS}. 
As shown, \textbf{FoSS} produces smooth and accurate trajectories across most scenarios. 
Even in challenging cases involving sharp turns or U-turns, the model outputs multiple high-quality predictions that closely follow lane geometry, demonstrating strong multimodal reasoning and adherence to road constraints.  
In frequent lane-change scenarios (Fig.~\ref{Figure 4}(f)), performance slightly degrades, the predicted trajectories exhibit minor jitter. 
This may be due to the limited ability to capture transient high-frequency motion cues. Lane boundaries are represented as low-frequency components during spectral decomposition, which may underrepresent rapid lateral maneuvers. 
Overall, these qualitative results confirm the model’s robustness and its capability to produce diverse and road-compliant future trajectories across complex traffic conditions.

\subsection{Ablation Study}
To quantify the contribution of each key component, we conduct an ablation study on the Argoverse~2 dataset, as summarized in Table~\ref{tab:argo2}. 
Removing the frequency-domain branch (F.D.B.) degrades performance notably, with minADE$_6$ increasing from 0.65 to 0.71 and minFDE$_6$ from 1.29 to 1.36, indicating that frequency cues are essential for capturing global motion trends and reducing noise in long-horizon prediction.  
Disabling the progressive helix reordering module (P.H.R.M.) causes a moderate drop (\textbf{+6.2\%} minADE, \textbf{+2.3\%} minFDE), showing that ordered spectral traversal enhances structural coherence and fine-grained motion extraction.  
When the Fourier S.S.M. is removed, performance declines to 0.70/1.35/0.17 on minADE/minFDE/MR, underscoring the selective SSM’s importance for stable spectral–temporal fusion.  
Replacing the cross-attention fusion with a simple Concat+MLP further increases errors (0.69/1.33/0.16), confirming that token-wise cross-state interaction effectively aligns temporal and frequency representations.  
Overall, the full model achieves the best results (minADE$_6$=0.65, minFDE$_6$=1.29, MR$_6$=0.15), demonstrating the complementary benefits of frequency–time dual-branch modeling.

\section{Conclusion and Future Work}
This work present \textbf{FoSS}, a dual-branch framework that combines frequency-domain decomposition and input-dependent state-space modeling to achieve an efficient yet expressive trajectory predictor. 
With cross-domain attention and adaptive trajectory fusion, the approach maintains robustness and multimodality under linear computational cost.
Extensive experiments on Argoverse~1 and~2 demonstrate consistent gains, improving minADE and minFDE by up to \textbf{14.8\%} and \textbf{22.5\%}, while reducing inference latency by \textbf{22\%} and parameter count by \textbf{40\%}. 
These results confirm the framework’s effectiveness in complex traffic scenarios. 
Future work will focus on long-tail corner cases involving abrupt low-to-high frequency motion changesto ensure robust forecasting in highly dynamic environments.

\section{Acknowledgement}
This work was supported in part by UKRI Innovate UK projects (10071278, 10099265), as well as UKRI under the Horizon Europe funding guarantee (EP/Y028031/1, EP/U536908/1), as part of the European Commission MSCA COVER (101086228) and ANT (101169439) projects. K. Wang would like to acknowledge the support by Royal Society Industry Fellow scheme (IF$\setminus$R2$\setminus$23200104). Y. Huang would like to acknowledge the support by Research England’s Enhancing Research Culture grant.

{
    \small
    \bibliographystyle{ieeenat_fullname}
    \bibliography{main}
}


\end{document}